\documentclass[runningheads]{styles/llncs}
\usepackage{graphicx}
\usepackage{comment}
\usepackage{amsmath,amssymb} 
\usepackage{color}
\usepackage{adjustbox}
\usepackage{subfig}
\usepackage{multirow}
\usepackage{amssymb}
\usepackage{pifont}
\usepackage{multirow, boldline}
\usepackage{ctable}
\usepackage{xcolor}
\usepackage[bottom]{footmisc}
\usepackage{listings}
\usepackage{wrapfig}
\usepackage{makecell}
\usepackage{hyperref}
\hypersetup{pagebackref=true,breaklinks=true,letterpaper=true,colorlinks,bookmarks=false}
\usepackage{breakcites}
\hypersetup{
     colorlinks   = true,
     citecolor    = green
}
\hypersetup{linkcolor=red}

\newcommand*\samethanks[1][\value{footnote}]{\footnotemark[#1]}

\newcommand{\etal}{\textit{et al}.}

\newcommand{\cmark}{\ding{51}}%
\definecolor{darkgreen}{rgb}{0.0, 0.6, 0.2}

\definecolor{MyRed}{rgb}{0.8,0.2,0}

\definecolor{MyBlue}{rgb}{0,0,1.0}

\definecolor{dkgreen}{rgb}{0,0.6,0}
\definecolor{gray}{rgb}{0.5,0.5,0.5}
\definecolor{mauve}{rgb}{0.58,0,0.82}

\begin{document}
\pagestyle{headings}
\mainmatter
\def\ECCVSubNumber{0000}  

\def\JL#1{{\color{red}JL: \it #1}}

\title{Learning Temporally Invariant and \\ Localizable Features via Data Augmentation \\ for Video Recognition}

\titlerunning{Temporally Invariant Data Augmentation for Video Recognition}
%
\author{Taeoh Kim\thanks{Equal contribution}\inst{1} \and
Hyeongmin Lee\samethanks\inst{1} \and
MyeongAh Cho\samethanks\inst{1} \and
Ho Seong Lee\inst{2} \and \\
Dong Heon Cho\inst{2} \and
Sangyoun Lee\inst{1}\thanks{Corresponding Author}}
\authorrunning{T. Kim et al}
%
\institute{Yonsei University, Seoul, South Korea \and
Cognex Deep Learning Lab, Seoul, South Korea \\
\email{\{kto, minimonia, maycho0305, syleee\}@yonsei.ac.kr} \\ \email{\{hoseong.lee, david.cho\}@cognex.com}}

\maketitle

\begin{abstract}

Deep-Learning-based video recognition has shown promising improvements along with the development of large-scale datasets and spatiotemporal network architectures.
In image recognition, learning spatially invariant features is a key factor in improving recognition performance and robustness. 
Data augmentation based on visual inductive priors, such as cropping, flipping, rotating, or photometric jittering, is a representative approach to achieve these features.
Recent state-of-the-art recognition solutions have relied on modern data augmentation strategies that exploit a mixture of augmentation operations.
In this study, we extend these strategies to the temporal dimension for videos to learn temporally invariant or temporally localizable features to cover temporal perturbations or complex actions in videos. 
Based on our novel temporal data augmentation algorithms, video recognition performances are improved using only a limited amount of training data compared to the spatial-only data augmentation algorithms, including the 1st Visual Inductive Priors (VIPriors) for data-efficient action recognition challenge.
Furthermore, learned features are temporally localizable that cannot be achieved using spatial augmentation algorithms. Our source code is available at \url{https://github.com/taeoh-kim/temporal_data_augmentation}.

\end{abstract}
\section{Introduction}

Many augmentation techniques have been proposed to increase the recognition performance and robustness for an environment with limited training data or to prevent overconfidence and overfitting of large-scale data, such as ImageNet~\cite{krizhevsky2012imagenet}. These techniques can be categorized into data-level augmentation~\cite{krizhevsky2012alexnet, vggnet, autoaugment, fastautoaugment, randaugment, augmix, cutout, hideandseek}, data-level mixing~\cite{mixup, cutmix, cutblur, attributemix, attentivecutmix, smoothmix}, and in-network augmentation~\cite{dropout, dropblock, stochasticdepth, shakeshake, shakedrop, regvideo, manimixup}.
Data augmentation is an important component for recent state-of-the-art self-supervised learning~\cite{moco, simclr, pirl}, semi-supervised learning~\cite{uda, mixmatch, remixmatch, fixmatch}, 
self-learning~\cite{noisystudent}, and generative models~\cite{crgan, diffauggan, bcrgan, dagan} because of its ability to learn invariant features.

The purpose of data augmentation in image recognition is to enhance the generalizability via learning spatially invariant features. Augmentation, such as geometric (cropping, flipping, rotating, \textit{etc.}) and photometric (brightness, contrast, color, \textit{etc.}) transformation, can model uncertain variances in a dataset. 
Recent algorithms have exhibited state-of-the-art performances in terms of the complexity-accuracy trade-off~\cite{fastautoaugment, randaugment} or robustness~\cite{robustness, augmix}. Some approaches~\cite{cutmix, cutblur} learn localizable features that can be used as transferable features for the localization-related tasks, such as object detection and image captioning. They simultaneously learn what to and where to focus for recognition. 

\begin{figure*}[!t]
	\centering
	\subfloat
	{\includegraphics[width=0.155\linewidth]{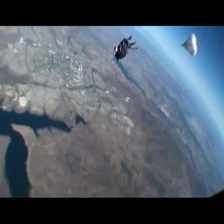}}\ 
	\subfloat
	{\includegraphics[width=0.155\linewidth]{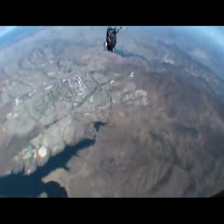}}\
	\subfloat
	{\includegraphics[width=0.155\linewidth]{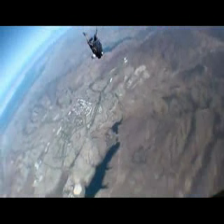}}\
	\hfill
	\subfloat
	{\includegraphics[width=0.155\linewidth]{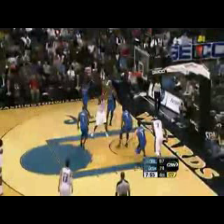}}\ 
	\subfloat
	{\includegraphics[width=0.155\linewidth]{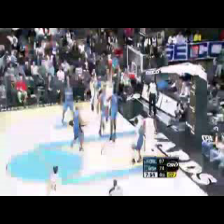}}\
	\subfloat
	{\includegraphics[width=0.155\linewidth]{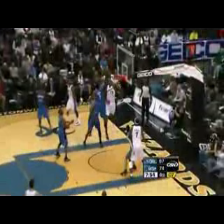}}\ \\
	\caption{Example clips of temporal perturbations. \textit{Left}: Geometric perturbation across frames in a sky-diving video due to extreme camera and object movement. \textit{Right}: Photometric perturbation across frames in a basketball stadium due to camera flashes.}
	\label{fig_perturbation}
\end{figure*}

Despite evolving through numerous algorithms in image recognition, exploration into data augmentation and regularization in video recognition has rarely been done. 
In videos, temporal variations and perturbations should be considered.
For example, Fig. \ref{fig_perturbation} depicts temporal perturbations across frames in a video. 
This perturbation can be a geometric perturbation, such as translation, rotation, scale, and so on, or a photometric perturbation, such as brightness, contrast, and so on. To handle perturbation, both well-studied spatial augmentation and temporally varying data augmentation should be considered.

In this paper, we propose several extensions for temporal robustness. More specifically, temporally invariant and localizable features can be modeled via data augmentations. 
In this paper, we extend upon two recent examples of well-studied spatial augmentation techniques: data-level augmentation and data-level mixing. To the best of our knowledge, this is the first study that deeply analyzes temporal perturbation modeling via data augmentation in video recognition.

The contributions of this paper can summarized as follows:

\begin{itemize}
	\item {We propose an extension of RandAugment~\cite{randaugment}, called RandAugment-T, to conduct data-level augmentation for video recognition. It can temporally model varying levels of augmentation operations.}
	\item {We also propose the temporal extensions of CutOut~\cite{cutout}, MixUp~\cite{mixup}, and CutMix~\cite{cutmix} as examples of deleting, blending, and cut-and-pasting data samples. Considering the temporal dimension improves recognition performance and the temporal localization abilities.}
	\item {The recognition results of the proposed extensions on the UCF-101~\cite{soomro2012ucf101} subset for the 1st Visual Inductive Priors (VIPriors) for data-efficient action recognition challenge, and the HMDB-51~\cite{kuehne2011hmdb} dataset exhibit performance  improvements compared to the spatial-only versions in a simple baseline.} 
\end{itemize}

\section{Related Works}

\subsection{Data augmentation}

\subsubsection{Data-level augmentation}

First, to enlarge the generalization performance of a dataset and to reduce the overfitting problem of preliminary networks, various data augmentation methods, such as rotate, flip, crop, color jitter~\cite{krizhevsky2012imagenet}, and scale jitter~\cite{vggnet} have been proposed.
CutOut~\cite{cutout} deletes a square-shaped box at a random location to encourage the network focus on various properties of images, to avoid relying on the most discriminative regions. Hide-and-Seek~\cite{hideandseek} is a similar approach, but it deletes multiple regions that are sampled from grid patches.

Recently, the methodology of combining more than one augmentation operation has been proposed. Cubuk~\etal~\cite{autoaugment} propose a reinforcement learning-based approach to search for the optimal data augmentation policy in the given dataset. 
However, because the search space is too large, it requires extensive time to determine the optimal policy. 
Although an approach to mitigate this problem has been proposed~\cite{fastautoaugment}, it is difficult hard and time-consuming to determine the optimal augmentation strategy. 
To solve this, Cubuk~\etal~\cite{randaugment} propose RandAugment, which randomly samples augment operations from the candidate list and cascades them. 
Similarly, Hendrycks~\etal~\cite{augmix} propose an approach called AugMix that parallelly blends images that have been augmented by the operations sampled from a set of candidates.  

These techniques can model uncertain spatial perturbation, such as the geometric transform, photometric transform, or both. Because studies have focused on static images, applying these approaches to videos is a straightforward extension. For videos, Ji~\etal~\cite{ji2019learning} propose temporal augmentation operations called time warping and time masking, which randomly adjust or skip temporal frames. In contrast, in this paper, we focus on the temporally varying augmentation.

\subsubsection{Data-level mixing}

Together with data augmentation algorithms, augmentation strategies using multiple samples have been proposed.
Zhang~\etal~\cite{mixup} propose an approach called MixUp to manipulate images with more than one image. This approach makes a new sample by blending two arbitrary images and interpolating their one-hot ground-truth labels. This encourages the model to behave linearly in-between training examples. 
CutMix~\cite{cutmix} combines the concepts of CutOut and MixUp, by taking the best of both worlds. 
It replaces a square-shaped deleted region in CutOut with a patch from another image. 
This encourages the model to learn not only what to recognize but also where to recognize it. 
It can be interpreted as spatially localizable feature learning.
Inspired by CutMix, several methods have been proposed. 
CutBlur~\cite{cutblur} propose a CutMix-like approach to solving the restoration problem by cut-and-pasting between low-resolution and high-resolution images. They also proposed CutMixUp, which combines MixUp and CutMix. CutMixUp blends the two images inside the one of the masks of CutMix to relax extreme changes in boundary pixels.
Attribute~Mix~\cite{attributemix} uses masks of any shape, not only square-shaped masks.
Attentive~CutMix~\cite{attentivecutmix} also discards the square-shaped masks. It uses multiple patches sampled from the grid and replaces the regions with another image. 
Smoothmix~\cite{smoothmix} focuses on the 'strong edge' problem caused by the boundary of the masks.

Although numerous data manipulation methods, including deleting, blending, and cut-and-pasting, have successfully augmented many image datasets, their ability when applied to video recognition to learn  temporally invariant and localizable features has not yet been explored. 

\subsubsection{In-network augmentation}

Apart from the data-level approaches, several studies have proposed in-network augmentation algorithms. 
These have usually involved the design of stochastic networks to undergo augmentation at the feature-level to reduce predictive variance and to learn more high-level augmented features rather than to learn features from low-level augmentations. Dropout~\cite{dropout} is the very first approach to regularize the overfitted models. Other approaches, such as DropBlock~\cite{dropblock}, Stochastic depth~\cite{stochasticdepth}, Shake-Shake~\cite{shakeshake}, and ShakeDrop~\cite{shakedrop} regularization, have been proposed. Manifold-MixUp~\cite{manimixup} propose a mixing strategy like MixUp but is used instead in the feature space. The most similar approach to this study is a regularization method for video recognition called Random Mean Scaling~\cite{regvideo}. It randomly adjusts spatiotemporal features in video networks. In contrast, our approach focuses on data-level manipulation and is extended from  spatial-only algorithms into the temporal worlds.

\subsection{Video recognition}

For video action recognition, like image recognition, various architectures have been proposed to capture spatiotemporal features from videos. 
In \cite{tran2015learning}, Tran \textit{et al.} proposed C3D, which extracts features containing objects, scenes, and action information through 3D convolutional layers and then simply passes them through a linear classifier. 
In \cite{tran2018closer}, a (2+1)D convolution that focuses on layer factorization rather than 3D convolution is proposed.
It is composed using a 2D spatial convolution followed by 1D temporal convolution. 
In addition, the non-local block~\cite{wang2018non} and GloRe~\cite{chen2019graph} modules have been suggested to capture long-range dependencies via self-attention and graph-based modules. 
By plugging them into 3D ConvNet, the network can learn long-distance relations in both space and time. 
Another approach is two-stream architecture~\cite{wang2016temporal, stroud2020d3d, ryoo2019assemblenet}. 
In \cite{carreira2017quo}, a two-stream 3D ConvNet inflated from the deep image classification network and pre-trained features is proposed and achieves state-of-the-art performance by pre-training with the Kinetics dataset, a large-scale action recognition dataset. 
Based on this architecture, Xie \textit{et al.} \cite{xie2017rethinking} combined a top-heavy model design, temporally separable convolution, and spatiotemporal feature-gating blocks to make low-cost and meaningful features. 
Recently, SlowFast~\cite{feichtenhofer2019slowfast} networks that consist of a slow path for semantic information and a fast path for rapidly changing motion information exhibit competitive performance with a different frame rate sampling strategy. 
In addition, RESOUND~\cite{li2018resound} proposed a method to reduce the static bias of the dataset, and an Octave convolution~\cite{chen2019drop} is proposed to reduce spatial redundancy by dividing the frequency of features. A debiasing loss function~\cite{choi2019can} is proposed to mitigate the strong scene bias of networks and focus on the actual action information.

Since the advent of the large-scale Kinetics dataset, most action recognition studies have pre-trained the backbone on Kinetics, which guarantees basic performance. 
However, based on the results of the study by \cite{hara2018can}, architectures with numerous parameters are significantly overfitted when learning from scratch on relatively small datasets, such as UCF-101~\cite{soomro2012ucf101} and HMDB-51~\cite{kuehne2011hmdb}. This indicates that training without a pre-trained backbone is a challenging issue. Compared to existing studies that have been focused on novel dataset and architectures, we focus on regularization techniques, such as data augmentation, to prevent overfitting via learning invariance and robustness in terms of spatiality and temporality.
\section{Methods}

\subsection{Data-level temporal data augmentations}

\begin{wrapfigure}{r}{0.5\linewidth}
	\vspace{-1.0cm}
	\begin{lstlisting}	
	def randaugment_T(X, N, M1, M2): 
	"""Generate a set of distortions.
	
	Args:
	X: Input video (T x H x W)
	N: Number of augmentation transformations 
	to apply sequentially.
	M1, M2: Magnitudes for both temporal ends.
	"""
	
	ops = np.random.choice(transforms, N)
	M = np.linspace(M1, M2, T)
	return [[op(X, M[t]) for t in range(T)] for op in ops]
	\end{lstlisting}
	\vspace{-0.5cm}
	\caption{\small{Pseudo-code for RandAugment-T based on Numpy in Python. Template is borrowed from~\cite{randaugment}}}
	\label{fig:randaugt}
	\vspace{-0.5cm}
\end{wrapfigure}

First, we extend the existing RandAugment~\cite{randaugment} method for video recognition. RandAugment has two hyper-parameters for optimization. One is the number of augmentation operations to apply, N, and the other is the magnitude of the operation, M. A grid search of these two parameters in a given dataset produces state-of-the-art performance in image recognition. 

\begin{figure*}[!t]
	\centering
	\subfloat
	{\includegraphics[width=0.8\linewidth]{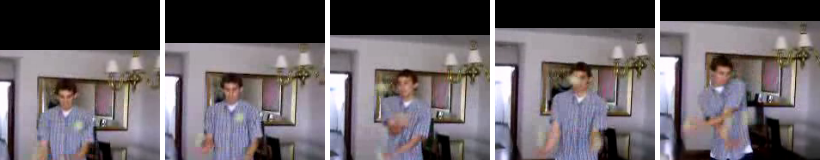}}\ \\[0.2ex]
	\subfloat[(a) Temporally varying geometric augmentations (Top: vertical-down translation, Bottom: clockwise rotation)]
	{\includegraphics[width=0.8\linewidth]{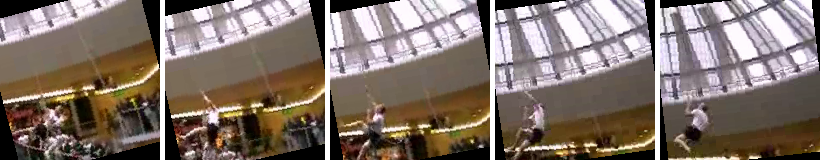}}\ \\
	\subfloat
	{\includegraphics[width=0.8\linewidth]{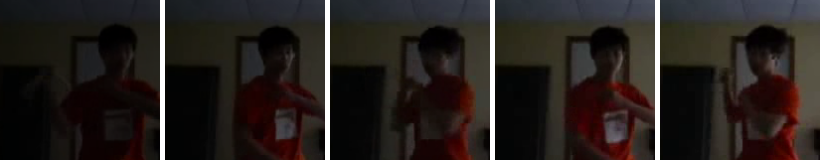}}\ \\[0.2ex]
	\subfloat[(b) Temporally varying photometric augmentations (Top: increasing brightness, Bottom: decreasing contrast)]
	{\includegraphics[width=0.8\linewidth]{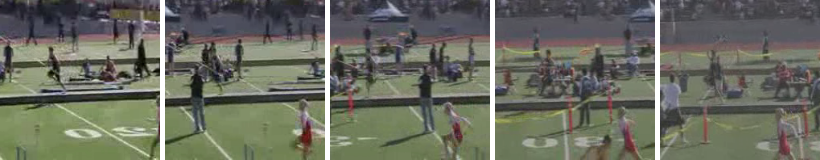}}\ \\
	\caption{Example of temporally varying data augmentation operations for RandAugment-T}
	\label{fig:taugexample}
\end{figure*}

For video recognition, RandAugment is directly applicable to every frame of a video; however, this limits temporal perturbation modeling. To cover temporally varying transformations, we propose RandAugment-T, which linearly interpolates between two magnitudes from the first frame to the last frame in a video clip. 
The pseudo-code for RandAugment-T is described in Fig.~\ref{fig:randaugt}. It receives three hyper-parameters: N, M1, and M2, where N is the number of operations, which is the same as RandAugment, and M1 and M2 indicate the magnitudes for both temporal ends, which can be any combination of levels. The set of augmentation operations (\texttt{transforms} in Fig.~\ref{fig:randaugt}) is identical to RandAugment. 
However, \texttt{rotate}, \texttt{shear-x}, \texttt{shear-y}, \texttt{translate-x}, and \texttt{translate-y} can model temporally varying geometric transformation, such as camera or object movements (Fig.~\ref{fig:taugexample}(a)), and \texttt{solarize},  \texttt{color}, \texttt{posterize}, \texttt{contrast}, \texttt{brightness}, and \texttt{sharpness} can model photometric transformation, such as brightness or contrast changes due to the auto-shot mode in a camera (Fig. ~\ref{fig:taugexample}(b)). The remaining operations (\texttt{identity}, \texttt{autocontrast}, and \texttt{equalize}) have no magnitudes that are applied evenly across frames.
	
\subsection{Data-level temporal deleting, blending, and cut-and-pasting}
\label{regularization}

\begin{figure*}[!t]
	\centering
	\subfloat
	{\includegraphics[width=0.49\linewidth]{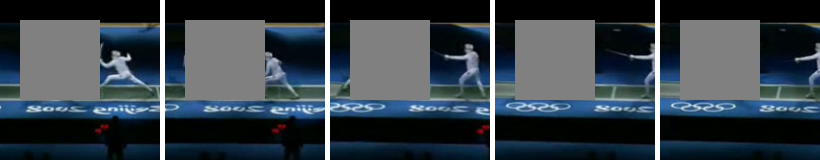}}\
	\hfill
	\subfloat
	{\includegraphics[width=0.49\linewidth]{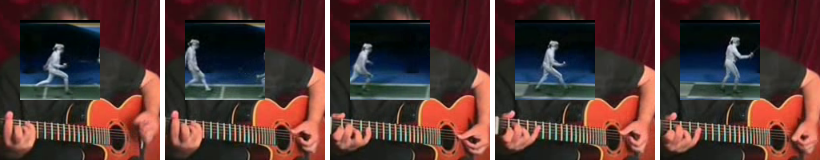}}\ \\[-2ex]
	\subfloat
	{\includegraphics[width=0.49\linewidth]{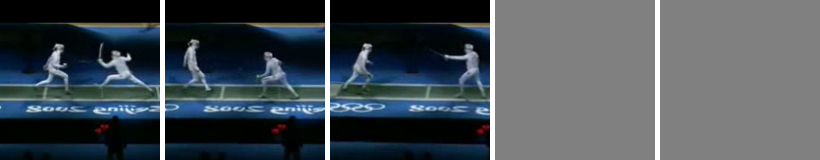}}\
	\hfill
	\subfloat
	{\includegraphics[width=0.49\linewidth]{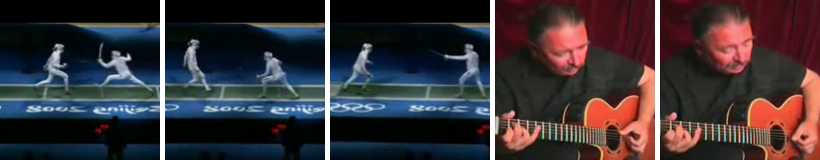}}\ \\[-2ex]
	\subfloat[\small{(a) \textit{Top}: CutOut~\cite{cutout}, \textit{Middle}: FrameCutOut, \textit{Bottom}: CubeCutOut}]
	{\includegraphics[width=0.49\linewidth]{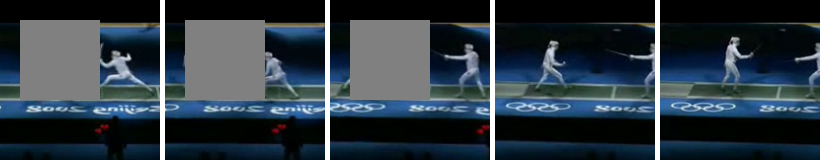}}\
	\hfill
	\subfloat[\small{(b) \textit{Top}: CutMix~\cite{cutmix}, \textit{Middle}: FrameCutMix, \textit{Bottom}: CubeCutMix}]
	{\includegraphics[width=0.49\linewidth]{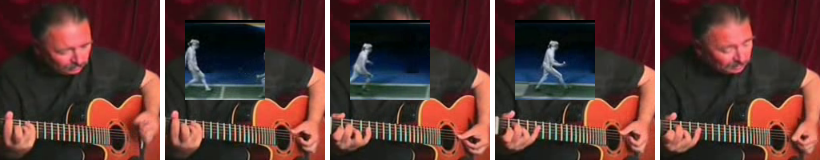}}\ \\[-2ex]
	\subfloat
	{\includegraphics[width=0.49\linewidth]{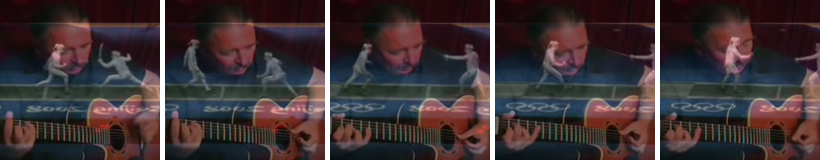}}\
	\hfill
	\subfloat
	{\includegraphics[width=0.49\linewidth]{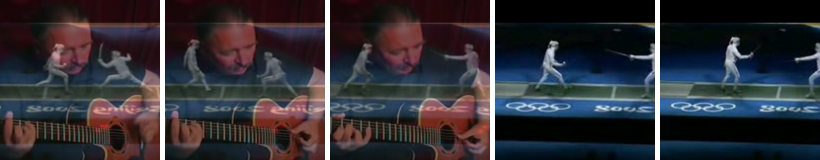}}\ \\[-2ex]
	\subfloat[\small{(c) \textit{Top}: MixUp~\cite{mixup}, \textit{Bottom}: CutMixUp~\cite{cutblur}}]
	{\includegraphics[width=0.49\linewidth]{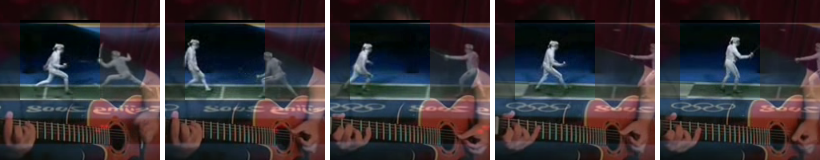}}\
	\hfill
	\subfloat[\small{(d) \textit{Top}: FrameCutMixUp, \textit{Bottom}: CubeCutMixUp}]
	{\includegraphics[width=0.49\linewidth]{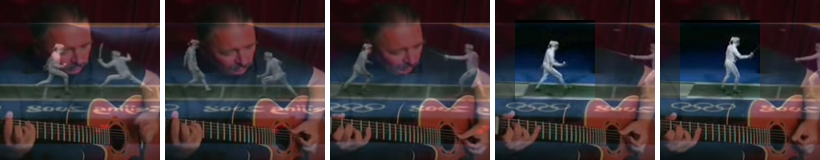}}\ \\[0.5ex]
	\subfloat[\small{(e) FadeMixUp}]
	{\includegraphics[width=0.49\linewidth]{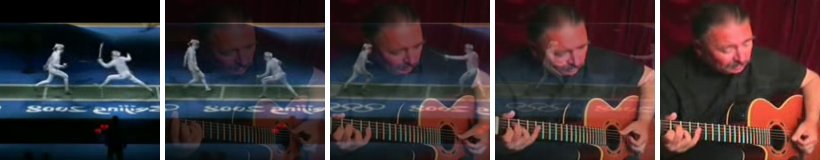}}\
	\hfill
	\caption{Visual comparison of data-level deleting, blending, and cut-and-pasting for videos. Desired ground-truth labels are calculated by the ratio of each class: \textit{Fencing} and \textit{PlayingGuitar}.}
	\label{fig_frameworkcomparison}
\end{figure*}

Regularization techniques, which have been proposed for image recognition, such as CutOut~\cite{cutout}, MixUp~\cite{mixup}, and CutMix~\cite{cutmix}, can be applied identically across frames in a video. CutMixUp is a combination of MixUp and CutMix, which is proposed in~\cite{cutblur}, can also be used for relaxing the unnatural boundary changes. 

In this section, we propose temporal extensions of the above algorithms. FrameCutOut and CubeCutOut are the temporal and spatiotemporal extensions of CutOut (Fig~\ref{fig_frameworkcomparison} (a)), respectively. CutOut encourages the network to better use the full context of the images, rather than relying on a small portion of specific spatial regions. Similarly, FrameCutOut encourages the network to better use the full temporal context and the full spatiotemporal context by CubeCutOut. 

FrameCutMix and CubeCutMix are extensions of CutMix~\cite{cutmix} (Fig~\ref{fig_frameworkcomparison} (b)). CutMix is designed for the learning of spatially localizable features. Cut-and-paste mixing between two images encourages the network to learn where to recognize features. Similarly, FrameCutMix and CubeCutMix are designed for the learning of temporally and spatiotemporally localizable features in a video. Like CutMix, the mixing ratio $\lambda$ is sampled from the beta distribution $Beta(\alpha, \alpha)$, where $\alpha$ is a hyper-parameter, and the locations for random frames or random spatiotemporal cubes are selected based on $\lambda$.

Like CutMixUp~\cite{cutblur}, which is the unified version of MixUp~\cite{mixup} and CutMix~\cite{cutmix}, FrameCutMixUp and CubeCutMixUp can be designed similarly (Fig~\ref{fig_frameworkcomparison} (c) and (d)) to relax extreme boundary changes between two samples. 
For these blend$+$cut-and-paste algorithms, MixUp is applied between two data samples by the mixing ratio $\lambda_1$, and the other hyper-parameter $\lambda_2$ is sampled from $Beta(2, 2)$. Based on $\lambda_2$, the region mask $\mathbf{M}$ is selected randomly similar to CutMix to cut-and-paste the MixUp-ed sample and one of the two original samples. The final mixed data and desired ground-truth labels are formulated as follows:
\begin{equation}
\begin{split}
\Tilde{x} =
\left\{
\begin{array}{ll}
(\lambda_1 x_A + (1-\lambda_1) x_B) \odot \mathbf{M} + x_A \odot (\mathbf{1} - \mathbf{M})  & \quad \mbox{if } \lambda_1 < 0.5 \\
(\lambda_1 x_A + (1-\lambda_1) x_B) \odot \mathbf{M} + x_B \odot (\mathbf{1} - \mathbf{M}) & \quad \mbox{if } \lambda_1 \geq 0.5
\end{array}
\right. \\ 
\Tilde{y} =
\left\{
\begin{array}{ll}
(\lambda_1 \lambda_2 + (1 - \lambda_1)) y_A + (1-\lambda_1) \lambda_2 y_B  & \quad \mbox{if } \lambda_1 < 0.5 \\
\lambda_1 \lambda_2 y_A + (1 - \lambda_1 \lambda_2) y_B & \quad \mbox{if } \lambda_1 \geq 0.5
\end{array}
\right.
\end{split}
\end{equation}
where $\Tilde{x}$, $\Tilde{y}$, and $\odot$ indicate the mixed data, modified label, and element-wise multiplication, respectively.

Finally, we propose another extension of MixUp, called FadeMixUp, inspired by the fade-in, fade-out, and dissolve overlap effects in videos. For FadeMixUp, in MixUp, the mixing ratio is smoothly changing along with temporal frames (Fig~\ref{fig_frameworkcomparison} (e)). 
In FadeMixUp, a list of the mixing ratios $\Tilde{\lambda}_t$ of a frame $t$ is calculated by linear interpolation between $\lambda - \gamma$ and $\lambda + \gamma$, where $\lambda$ is the mixing ratio of MixUp, and the $\gamma$ is sampled from $Uniform(0, min(\lambda, 1-\lambda))$. Because the adjustments in the mixing ratio at both ends are symmetric, the label is the same as MixUp.
\begin{equation}
\begin{split}
\Tilde{x_t} & =  \Tilde{\lambda_t} X_{A_t} + (1-\Tilde{\lambda}_t) X_{B_t} \\
\Tilde{y} & =  \lambda y_A + (1-\lambda) y_B, \\
\end{split}
\label{eq:fademixup}
\end{equation}

FadeMixUp can be modeled for temporal variations and can learn temporally localizable feature without sharp boundary changes, like other cut-and-pasting algorithms. Because many videos include these overlapping effects at the scene change, FadeMixUp can be applied naturally.

A summary of deleting, blending, and cut-and-pasting data augmentation algorithms is described in Table~\ref{tb:mixcomp}. In the table, a checkmark indicates the elements (pixels) that can be changed along the spatial or temporal axis via augmentation methods. Compared to the existing algorithms~\cite{cutout, cutmix, mixup, cutblur}, our proposed methods are extended temporally and spatiotemporally.

\begin{table}[!t]
	\centering
	\caption{\small{Comparison between deleting, blending, and cut-and-pasting frameworks.}}
	\resizebox{1.0\linewidth}{!}{
		\begin{tabular}{ll|ccc|ccc|cc|ccc}
			\toprule 
			& Type & \multicolumn{3}{c|}{Delete} & \multicolumn{3}{c|}{Cut-and-paste} & \multicolumn{2}{c|}{Blend} & \multicolumn{3}{c}{Blend $+$ Cut-and-paste} \\
			\cmidrule{2-13}
			& Name & \makecell{CutOut \\ \cite{cutout}} & \makecell{Frame \\ CutOut} & \makecell{Cube\\CutOut} & \makecell{CutMix \\ \cite{cutmix}} & \makecell{Frame\\CutMix} & \makecell{Cube\\CutMix} & \makecell{MixUp \\ \cite{mixup}} & \makecell{Fade\\MixUp} & \makecell{CutMixUp\\ \cite{cutblur}} & \makecell{Frame\\CutMixUp} & \makecell{Cube\\CutMixUp} \\
			\midrule
			Axis & Spatial & \cmark & & \cmark & \cmark & & \cmark & & & \cmark & & \cmark \\
			 & Temporal &  & \cmark & \cmark &  & \cmark & \cmark & &  \cmark &  & \cmark & \cmark \\
			\bottomrule
	\end{tabular}}
	\label{tb:mixcomp}
\end{table}

\section{Experiments}

\subsection{Experimental Settings}

For video action recognition, we train and evaluate the proposed method on the UCF-101~\cite{soomro2012ucf101} and HMDB-51~\cite{kuehne2011hmdb} datasets.
The UCF-101 dataset originally consists of 13,320 videos with 101 classes. The dataset consists of three training/testing splits, but we used the modified split provided by the 1st VIPriors action recognition challenge that consists of 4,795 training videos and 4,742 validation videos.
The HMDB-51 dataset consists of 6,766 videos with 51 classes. We use the original three training/testing splits for training and evaluation.

Our experiments are trained and evaluated on a single GTX 1080-ti GPU and are implemented using the PyTorch framework.
We use SlowFast-50~\cite{feichtenhofer2019slowfast} as the backbone network with 64 temporal frames because it is more lightweight and faster than other networks such as C3D~\cite{tran2015learning}, I3D~\cite{carreira2017quo}, and S3D~\cite{xie2017rethinking}, without any pre-training and optical-flow.
For the baseline, basic data augmentation, such as random crop with a size of 160, random scale jittering between [160, 200] for the short side of a video, and random horizontal flip, are applied. 
For optimization, the batch size is set to 16, the learning rate is set to 1e-4, and a weight decay of 1e-5 is used. Moreover, we incorporate the learning rate warm-up~\cite{cosinewarmup} and cosine learning rate scheduling~\cite{cosinelr} with the Adam optimizer~\cite{adam}. We train all models for 150 epochs.
For evaluation, we sample 10 clips uniformly along the temporal axis and average softmax predictions. For the challenge, following \cite{feichtenhofer2019slowfast}, we sample 30 clips.

\subsection{Data-level temporal data augmentations}

Table \ref{table:taugres} presents the recognition results on the UCF-101 validation set for the VIPriors challenge. For all result tables, \textbf{boldface} indicates the best results, and an \underline{underline} indicates the second best. RandAugment-spatial indicates an original implementation without temporal variations. In the temporal version, M1 of Fig. \ref{fig:randaugt} is sampled from $Uniform(0.1, M2)$, and M2 is set to M of the spatial RandAugment. For temporal$+$, M1 and M2 are set to M$-\delta$ and M$+\delta$, respectively, where $\delta$ is sampled from $Uniform(0, 0.5\times M)$. 
For Mix in Table \ref{table:taugres}, it randomly chooses the spatial or temporal$+$ variations. The results reveal that solely applying RandAugment drastically improves recognition performance. Among them, temporally expanded RandAugment-T (temporal$+$) exhibits the best performance. For all RandAugment results, to produce the best accuracy, a grid search of two hyper-parameters: N $\in[1, 2, 3]$ and M $\in[3, 5, 10]$, is used.

\begin{table}[!t]
	\setlength{\tabcolsep}{3pt}
	
	\centering
	\begin{minipage}{.5\linewidth} 
		\centering
		\caption{\small{Data Augmentation Results}}
		\label{table:taugres}
		\begin{adjustbox}{width=1.0\linewidth}
			\begin{tabular}{l|l|cc}
				\toprule
				& Range &  Top-1 Acc. & Top-5 Acc. \\
				\midrule
				Baseline &  &  49.37 & 73.62 \\
				RandAugment & Spatial &  66.87 & 88.04 \\
				& Temporal &  67.33 & 88.42 \\
				& Temporal+ &  \textbf{69.23} & \textbf{89.20} \\
				& Mix &  \underline{68.24} & \underline{89.25} \\
			\end{tabular}
		\end{adjustbox}
	\end{minipage} \quad%
	\begin{minipage}{.4\linewidth}
		\centering
		\caption{\small{Data Deleting Results}}
		\label{table:toutres}
		\begin{adjustbox}{width=1.0\linewidth}
			\begin{tabular}{l|cc}
				\toprule
				&  Top-1 Acc. & Top-5 Acc. \\
				
				\midrule
				
				Baseline &  \textbf{49.37} & \textbf{73.62} \\
				CutOut &  46.01 & 69.80 \\
				FrameCutOut & \underline{47.60}  & 71.32 \\
				CubeCutOut & 47.45 & \underline{72.06} \\
			\end{tabular}
		\end{adjustbox}
	\end{minipage}%
	\vspace{-0.4cm}
	
\end{table}

\begin{table}[!t]
	\setlength{\tabcolsep}{3pt}
	
	\centering
	\begin{minipage}{.46\linewidth}
		\centering
		\caption{\small{Data Cut-and-paste Results}}
		\label{table:tmixres}
		\begin{adjustbox}{width=1.0\linewidth}
			\begin{tabular}{l|cc}
				\toprule
				&  Top-1 Acc. & Top-5 Acc. \\
				
				\midrule
				
				Baseline &  49.37 & 73.62 \\
				CutMix($\alpha=2$) &  50.81 & \underline{75.62} \\
				FrameCutMix($\alpha=2$) & 51.29  & 74.99 \\
				FrameCutMix($\alpha=5$) & \textbf{53.10}  & \textbf{76.61} \\
				CubeCutMix($\alpha=2$) & \underline{51.86} & 74.34 \\
				CubeCutMix($\alpha=5$) & 51.81 & 75.16 \\
			\end{tabular}
		\end{adjustbox}
	\end{minipage} \quad \quad
	\begin{minipage}{.4\linewidth}
		\centering
		\caption{\small{Data Blending Results}}
		\label{table:tblendres}
		\begin{adjustbox}{width=1.0\linewidth}
			\begin{tabular}{l|cc}
				\toprule
				&  Top-1 Acc. & Top-5 Acc. \\
				
				\midrule
				
				Baseline &  49.37 & 73.62 \\
				MixUp &  59.60 & \underline{82.56} \\
				FadeMixUp & 59.22 & 82.24 \\
				\midrule
				CutMixUp & 59.35 & 81.99 \\
				FrameMixUp & \textbf{60.67} & \textbf{83.47} \\
				CubeMixUp & \underline{59.85} & 82.20 \\
			\end{tabular}
		\end{adjustbox}
	\end{minipage} \quad%
	\vspace{-0.4cm}
	
\end{table}

\subsection{Data-level temporal deleting, cut-and-pasting, and blending}

The results of deleting data (CutOut, FrameCutOut, and CubeCutOut) are described in Table \ref{table:toutres}. 
For CutOut, an $80\times 80$ spatial patch is randomly deleted, and for FrameCutOut, 16 frames are randomly deleted. For CubeCutOut, an $80\times 80\times 16$ cube is randomly deleted. The results reveal that deleting patches, frames, or spatiotemporal cubes reduces recognition performance in a limited number of training datasets. Among them, CutOut exhibits the worst performance. 

For data cut-and-pasting, like that of CutMix~\cite{cutmix} and its extensions, the results are described in Table \ref{table:tmixres}. We apply the mixing probability of 0.5 for all methods and employ different hyper-parameters $\alpha$. Because the object size in the action recognition dataset is smaller than that in ImageNet~\cite{krizhevsky2012imagenet}, the mixing ratio should be sampled in a region close to 0.5 by sampling the large $\alpha$ in the beta distribution. The results demonstrate that the temporal and spatiotemporal extensions outperform the spatial-only mixing strategy. Because the probability of object occlusion during temporal mixing is lower than during spatial mixing, the performance of FrameCutMix is the most improved.

Finally, for data blending, compared to MixUp~\cite{mixmatch} and CutMixUp~\cite{cutblur}, the temporal and spatiotemporal extensions show slightly superior performance, which is described in Table \ref{table:tblendres}. Compared to deleting and cut-and-pasting augmentations, blending presents the best performances. Because the number of training data is limited, a linear convex combination of samples easily and effectively augments the sample space.

\begin{table}[!t]
	\centering
	\caption{\small{Temporal Augmentation Results on HMDB51 Dataset}}
	\resizebox{1.0\linewidth}{!}{
		\begin{tabular}{l|cc|cc|cc|cc}
			\toprule 
			& \multicolumn{2}{c}{Split-1} & \multicolumn{2}{c}{Split-2} & \multicolumn{2}{c}{Split-3} & \multicolumn{2}{c}{Average}\\
			\cmidrule{2-9}
			& Top-1 Acc. & Top-5 Acc.& Top-1 Acc. & Top-5 Acc. & Top-1  Acc. & Top-5 Acc. & Top-1 Acc. & Top-5 Acc. \\ \midrule	
			Baseline & 36.60 & 67.25 & 37.19 & 65.75 & 32.88 & 65.82 & 35.56 & 66.27 \\
			\midrule
			RandAug & \underline{47.45} & \underline{79.21} & \underline{47.12} & \underline{76.86} & \underline{47.45} & \underline{77.97} & \underline{47.34} & \underline{78.01} \\
			RandAug-T & \textbf{48.17} & \textbf{79.35} & \textbf{47.84} & \textbf{77.00} & \textbf{48.37} & \textbf{78.17} & \textbf{48.13} & \textbf{78.17} \\
			\midrule
			CutOut & \textbf{34.71} & \textbf{65.49} & \textbf{32.35} & 63.79 & \underline{31.76} & \underline{62.94} & \textbf{32.94 }& \textbf{64.07} \\
			FrameCutOut & 31.05 & 61.57 & \underline{32.16} & \textbf{65.36} & \textbf{31.87} & \textbf{64.18} & 31.69 & \underline{63.70} \\
			CubeCutOut & \underline{33.01} & \underline{63.99} & 32.04 & \underline{64.25} & 30.59 & 62.81 & \underline{31.88} & 63.68 \\
			\midrule
			CutMix & 33.95 & 64.27 & 33.69 & \underline{66.84} & 31.24 & \underline{63.53} & 32.96 & 64.88 \\
			FrameCutMix & \underline{34.97} & \textbf{65.56} & \underline{34.84} & \textbf{67.91} & \underline{33.27} & \underline{63.53} & \underline{34.36} & \underline{65.67} \\
			CubeCutMix & \textbf{35.10} & \underline{65.10} & \textbf{35.95} & 65.62 & \textbf{36.54} & \textbf{67.97} & \textbf{35.86} & \textbf{66.23} \\
			\midrule
			MixUp & 38.95 & 68.10 & \textbf{40.72} & 70.92 & \underline{40.20} & 71.31 & 39.96 & 70.11 \\
			CutMixUp &\textbf{ 40.92} & \textbf{71.07} &40.16 & 71.55 & 39.28 & \underline{71.48} & \underline{40.12} & \underline{71.37} \\
			FrameMixUp & 40.33 & \underline{70.98} & 40.52 & 70.85 & 39.02 & 70.65 & 39.96 & 70.83 \\
			CubeMixUp & \underline{40.72} & 70.65 & \underline{40.70} & \textbf{72.88} & \textbf{40.92} & \textbf{71.83} & \textbf{40.78} & \textbf{71.79} \\
			FadeMixUp & 39.80 & 70.39 & 40.46 & \underline{71.70} & 39.61 & 70.00 & 39.96 & 70.70 \\	
			\bottomrule	
	\end{tabular}}
	\label{tb:hmdb51}
\end{table}

\begin{table}[!t]
	\centering
	\caption{\small{Model Evaluation for VIPriors Challenge}}
	\resizebox{0.85\linewidth}{!}{
		\begin{tabular}{cc|c|c|c|c|cc}
			\toprule
			& Train Data  & Test Data & Augmentation & Regularization & Others & Top-1 Acc.  & Top-5 Acc.  \\ \midrule	
			& Train  & Val & & & & 49.37 & 73.62 \\
			\midrule
			& Train  & Val & & FrameMixUp & & 60.67 & 83.47 \\
			& Train  & Val & RandAug & & & 66.87 & 88.04 \\
			& Train  & Val & RandAug-T & & & \underline{69.23} & 89.20 \\
			& Train  & Val & RandAug-T & FadeMixUp & & 68.73 & \underline{89.27}  \\
			& Train  & Val & RandAug-T & FrameMixUp & & \textbf{69.70} & \textbf{89.84} \\
			\midrule
			& Train+Val  & Test & & & & 68.99 & - \\
			& Train+Val  & Test & RandAug-T & &  & 81.43 & - \\
			& Train+Val  & Test & RandAug-T & FadeMixUp &  & \underline{82.16} & - \\
			& Train+Val  & Test & RandAug-T & All Methods & Ensemble & \textbf{86.04} & - \\ \bottomrule	
	\end{tabular}}
	\label{tb:challenge}
\end{table}

\begin{table}[!t]
	\centering
	\caption{\small{Comparison between Entries of VIPriors Challenge}}
	\resizebox{0.75\linewidth}{!}{
		\begin{tabular}{cc|c|c|c|c}
			\toprule
			& Entry & Backbone  & Two-stream & Ensemble & Top-1 Acc.  \\ \midrule	
			& 1st place team & I3D, C3D, 3D-ResNet, R(2+1)D & \cmark & Across Model & \textbf{90.8}  \\
			& 2nd place team~\cite{chen2020viprior}  & TCDC & \cmark & Within Model & \underline{88.3}  \\
			& 3rd place team~\cite{luo2020viprior}  & SlowFast50, TSM & \cmark & Across Model & 87.6  \\
			\midrule
			& Ours  & SlowFast50 & &  & 82.2  \\
			& Ours  & SlowFast50 & & Within Model & 86.0  \\ \bottomrule	
	\end{tabular}}
	\label{tb:challenge_entry}
\end{table}

\subsection{Results on HMDB-51 dataset}

To determine the generalization to other datasets, we train and evaluate using the HMDB-51 dataset with its original splits. Generally, the recognition performance in HMDB-51 is inferior to the performance of UCF-101 due to its limited number of training samples. We use the same model and hyper-parameters as in UCF-101.

The results in Table~\ref{tb:hmdb51} indicate that the temporal extensions generally outperforms spatial-only versions, and similar to UCF-101, the RandAugment and blending demonstrate the best accuracy.

\begin{figure*}[!t]
	\centering
	\subfloat[\small{(a) Sample clip A: \textit{FrisbeeCatch}}]
	{\includegraphics[width=0.495\linewidth]{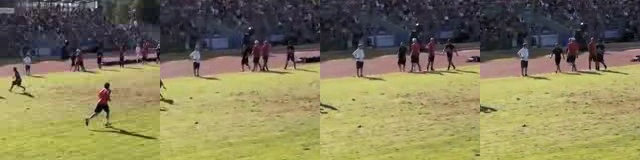}}\
	\hfill
	\subfloat[\small{(b) Sample clip B: \textit{JugglingBalls}}]
	{\includegraphics[width=0.495\linewidth]{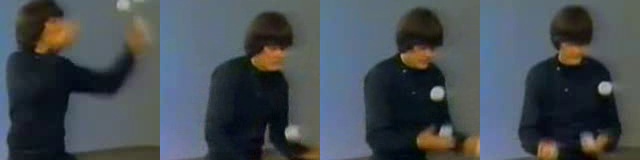}}\ \\[-2ex]
	\subfloat[\small{(c) MixUp-ed Clip}]
	{\includegraphics[width=0.495\linewidth]{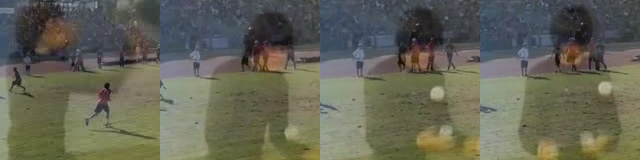}}\
	\hfill
	\subfloat[\small{(d) FadeMixUp-ed Clip}]
	{\includegraphics[width=0.495\linewidth]{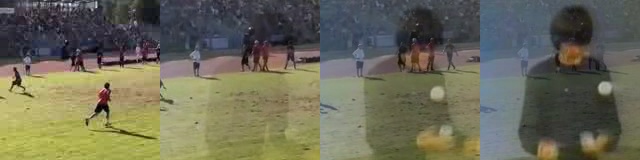}}\ \\[-2ex]
	\subfloat[\small{(e) CAM for \textit{FrisbeeCatch} on (c)}]
	{\includegraphics[width=0.495\linewidth]{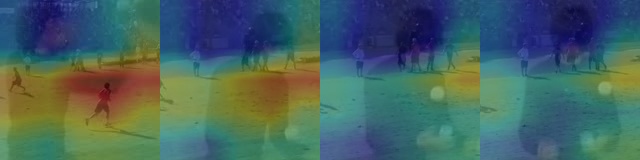}}\
	\hfill
	\subfloat[\small{(f) CAM for \textit{FrisbeeCatch} on (d)}]
	{\includegraphics[width=0.495\linewidth]{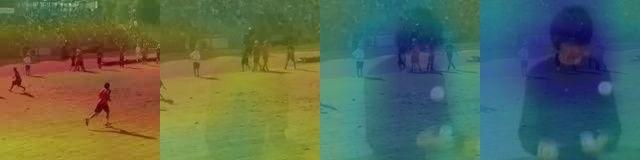}}\ \\[-2ex]
	\subfloat[\small{(g) CAM for \textit{JugglingBalls} on (c)}]
	{\includegraphics[width=0.495\linewidth]{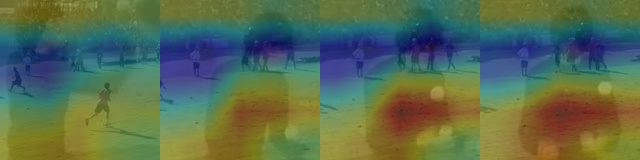}}\
	\hfill
	\subfloat[\small{(h) CAM for \textit{JugglingBalls} on (d)}]
	{\includegraphics[width=0.495\linewidth]{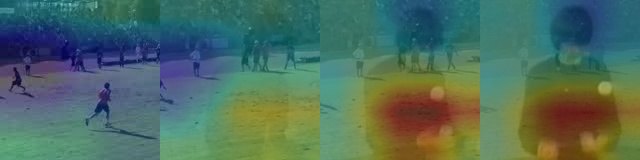}}\ \\[-2ex]
	\subfloat[\small{(i) CAM for \textit{FrisbeeCatch} on (a)}]
	{\includegraphics[width=0.495\linewidth]{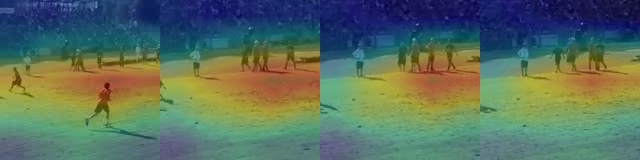}}\
	\hfill
	\subfloat[\small{(j) CAM for \textit{FrisbeeCatch} on (a)}]
	{\includegraphics[width=0.495\linewidth]{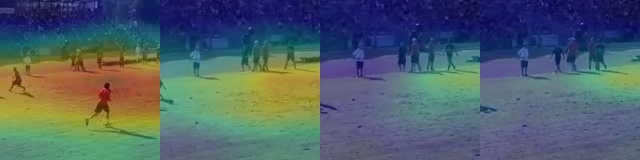}}\
	\caption{Class activation maps. \textit{Left}: MixUp, \textit{Right}: FadeMixUp}
	\label{fig_camforblend}
\end{figure*}

\subsection{1st VIPriors action recognition challenge}

Based on the comprehensive experimental results, we attend the 1st VIPriors action recognition challenge. In this challenge, any pre-training and external datasets are not allowed. 
The performance of various models is described in Table~\ref{tb:challenge}. 
For validation, applying both RandAugment-T and FrameMixUp perform the best. 
For the test set, 3,783 videos are provided without ground truths. 
Therefore, we report the results based on the challenge leaderboard. 
A combination of training and validation datasets including 9,537 videos are used to train the final challenge entries. 
According to the baseline accuracy of 68.99\%, adapting RandAugment-T improves the performance by only up to 81.43\%. Finally, we submitted an ensembled version of the different models that are trained using RandAugment-T and various mixing augmentation, to produce 86.04\% top-1 accuracy. 
The results including other challenge entries are described in Table~\ref{tb:challenge_entry}. The 1st place team proposes two-stream multi-scale spatiotemporal fusion strategy based on hand-craft optical flow and various 3D-ConvNets. The 2nd place team~\cite{chen2020viprior} also propose two-stream networks called 3D Temporal Central Difference Convolution (TCDC) based on C3D backbone. The 3rd place team~\cite{luo2020viprior} combines SlowFast network and Temporal Shift Module (TSM)~\cite{lin2019tsm} with two-stream networks. Compared to these methods, even if our final challenge results are inferior to them, our framework is much simple and comparative without using any two-stream strategy and model ensemble.

\subsection{Discussions}

\subsubsection{Why are the improvements not large?}

Although temporal extensions generally outperform spatial-only versions in data augmentation algorithms, performance improvements might be not large enough. The possible reasons for this are three-fold. The first reason is the lack of sufficient training data. The second is the lack of temporal perturbation, and the third is that datasets used for experiments consist of trimmed videos. 
Both UCF-101 and HMDB-51 datasets have little temporal perturbations. 
Therefore, applying spatial augmentation is sufficient to learn the context. Furthermore, both datasets are trimmed to have few temporal occlusions; therefore, no room is left to learn the ability to localize temporally. 
Compared to the image dataset, because the action region is relatively small, removing the spatial region can hurt the basic recognition performance for deleting and cut-and-pasting if the volume of training data is not adequate. In contrast, for blending, although it is an unnatural image, as said in~\cite{cutmix}, the blending can the exploit full region of frames. Therefore, it produces reasonable performance improvements.

\begin{figure*}[!t]
	\centering
	\subfloat[\small{(a) Sample clip A: \textit{Swing}}]
	{\includegraphics[width=0.495\linewidth]{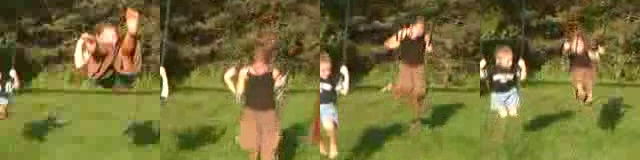}}\
	\hfill
	\subfloat[\small{(b) Sample clip B: \textit{Basketball}}]
	{\includegraphics[width=0.495\linewidth]{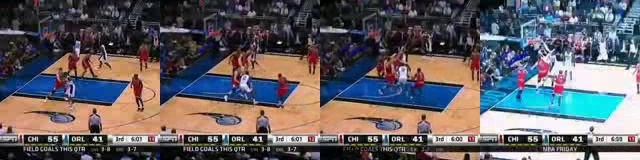}}\ \\[-2ex]
	\subfloat
	{\includegraphics[width=0.24\linewidth]{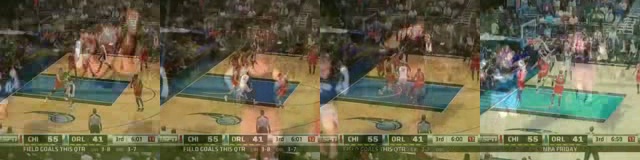}}\
	\hfill
	\subfloat
	{\includegraphics[width=0.24\linewidth]{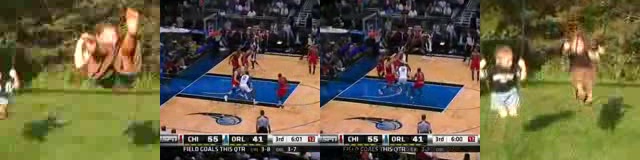}}\ 
	\hfill
	\subfloat
	{\includegraphics[width=0.24\linewidth]{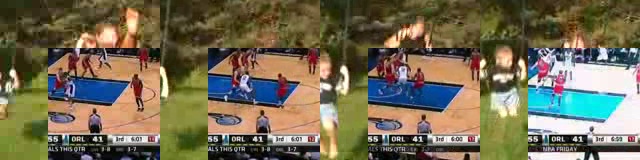}}\
	\hfill
	\subfloat
	{\includegraphics[width=0.24\linewidth]{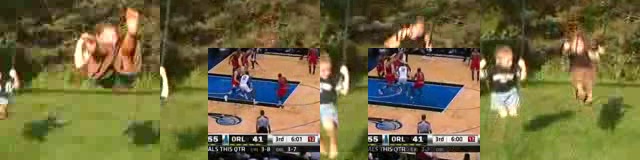}}\ \\[-2ex]
	\subfloat
	{\includegraphics[width=0.24\linewidth]{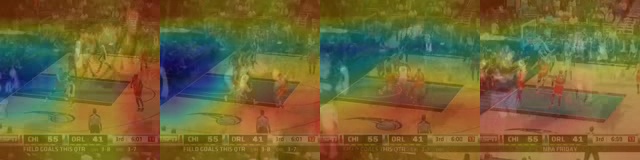}}\
	\hfill
	\subfloat
	{\includegraphics[width=0.24\linewidth]{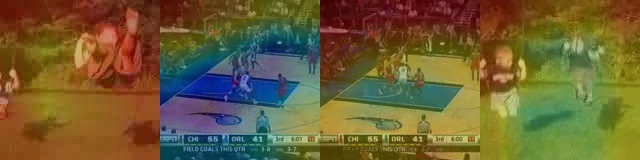}}\ 
	\hfill
	\subfloat
	{\includegraphics[width=0.24\linewidth]{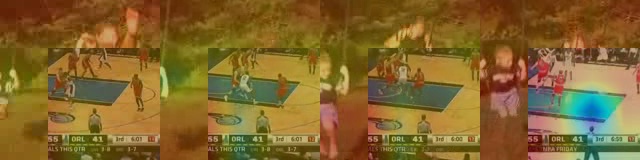}}\
	\hfill
	\subfloat
	{\includegraphics[width=0.24\linewidth]{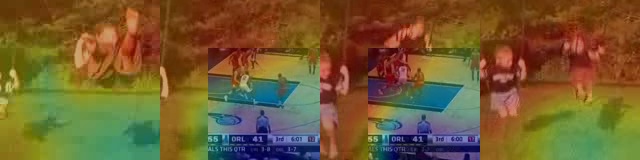}}\ \\[-2ex]
		\subfloat
	{\includegraphics[width=0.24\linewidth]{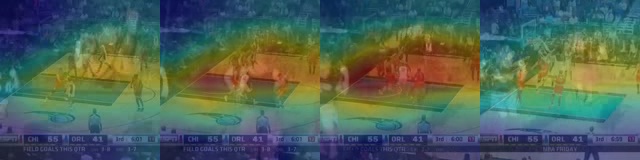}}\
	\hfill
	\subfloat
	{\includegraphics[width=0.24\linewidth]{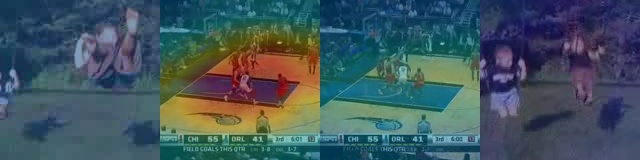}}\ 
	\hfill
	\subfloat
	{\includegraphics[width=0.24\linewidth]{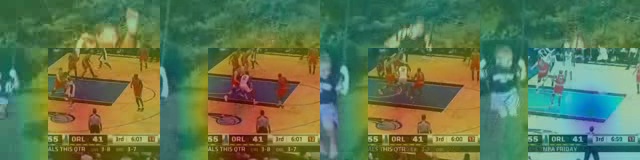}}\
	\hfill
	\subfloat
	{\includegraphics[width=0.24\linewidth]{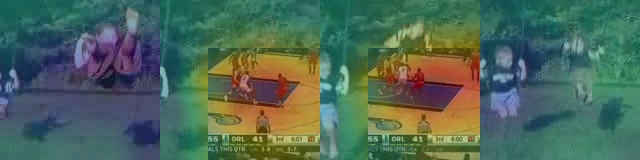}}\ \\[-2ex]
	\subfloat[\small{(c) MixUp}]
	{\includegraphics[width=0.24\linewidth]{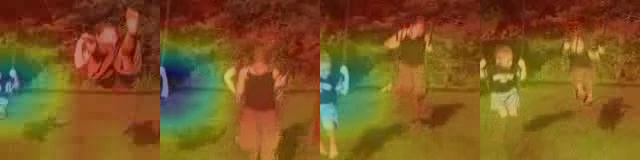}}\
	\hfill
	\subfloat[\small{(d) FrameCutMix}]
	{\includegraphics[width=0.24\linewidth]{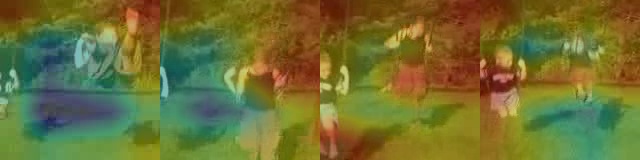}}\ 
	\hfill
	\subfloat[\small{(d) CutMix}]
	{\includegraphics[width=0.24\linewidth]{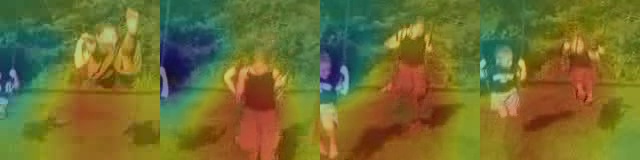}}\
	\hfill
	\subfloat[\small{(d) CubeCutMix}]
	{\includegraphics[width=0.24\linewidth]{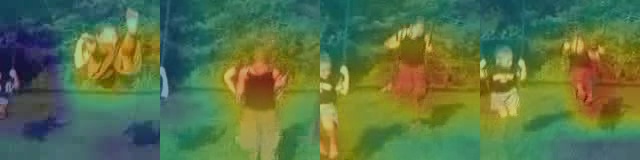}}\
	\caption{Class actionvation maps. For (c)-(f), from the top to the bottom row: mixed clips, CAMs for {\textit{Swing}}, CAMs for {\textit{Basketball}}, and CAMs for {\textit{Swing}} on pure clip (a), respectively.}
	\label{fig_camforstloc}
\end{figure*}

\subsubsection{Spatiotemporal class activation map visualization}

We visualize the learned features using the class activation map~\cite{cam} in Fig.~\ref{fig_camforblend}. In the SlowFast network, we use the features of the last convolutional layer in SlowPath. Fig.~\ref{fig_camforblend} (a) and (b) present example clips. Fig.~\ref{fig_camforblend} (c) and (d) are the visualizations of the clips using MixUp-ed and FadeMixUp-ed, respectively. 
In Fig.~\ref{fig_camforblend} (f) and (h) compared to Fig.~\ref{fig_camforblend} (e) and (g), the features of FadeMixUp are more localized temporally than those of MixUp. In Fig.~\ref{fig_camforblend} (j) compared to Fig.~\ref{fig_camforblend} (i), the activations of FadeMixUp are spatiotemporally localized better than those of MixUp in pure Clip A.

Fig.~\ref{fig_camforstloc} compares the spatiotemporal localization abilities of MixUp, CutMix, FrameCutMix, and CubeCutMix. Compared to MixUp, as stated in their paper~\cite{cutmix}, CutMix can spatially localize a basketball court or a person on a swing. However, compared to CubeCutMix, the activations of CutMix are not well localized temporally. FrameCutMix also cannot localize features like MixUp, but it can separate the weights of activation separately on the temporal axis.
\section{Conclusion}

In this paper, we proposed several extensions of data-level augmentation and data-level deleting, blending, and cut-and-pasting augmentation algorithms from the spatial (image) domain into the temporal and spatiotemporal (video) domain. 
Although applying spatial data augmentation increases the recognition performance in a limited amount of training data, extending temporal and spatiotemporal data augmentation boosts performance.
Moreover, our models that are trained on temporal augmentation can achieve temporal and spatiotemporal localization ability that cannot be achieved by the model trained only on spatial augmentation.
Our next step is an extension to a large-scale dataset, such as Kinetics~\cite{carreira2017quo}, or untrimmed videos.

\section*{Acknowledgments}

This research was supported by R\&D program for Advanced Integrated-intelligence for Identification (AIID) through the National Research Foundation of KOREA (NRF) funded by Ministry of Science and ICT (NRF-2018M3E3A1057289).

\clearpage
%
%
\bibliographystyle{utils/splncs04}
\bibliography{egbib}

\begin{thebibliography}{10}
\providecommand{\url}[1]{\texttt{#1}}
\providecommand{\urlprefix}{URL }
\providecommand{\doi}[1]{https://doi.org/#1}

\bibitem{remixmatch}
Berthelot, D., Carlini, N., Cubuk, E.D., Kurakin, A., Sohn, K., Zhang, H.,
  Raffel, C.: Remixmatch: Semi-supervised learning with distribution matching
  and augmentation anchoring. In: International Conference on Learning
  Representations (2019)

\bibitem{mixmatch}
Berthelot, D., Carlini, N., Goodfellow, I., Papernot, N., Oliver, A., Raffel,
  C.A.: Mixmatch: A holistic approach to semi-supervised learning. In: Advances
  in Neural Information Processing Systems. pp. 5049--5059 (2019)

\bibitem{carreira2017quo}
Carreira, J., Zisserman, A.: Quo vadis, action recognition? a new model and the
  kinetics dataset. In: proceedings of the IEEE Conference on Computer Vision
  and Pattern Recognition. pp. 6299--6308 (2017)

\bibitem{chen2020viprior}
Chen, H., Yu, Z., Liu, X., Peng, W., Lee, Y., Zhao, G.: 2nd place scheme on
  action recognition track of eccv 2020 vipriors challenges: An efficient
  optical flow stream guided framework. arXiv preprint arXiv:2008.03996  (2020)

\bibitem{simclr}
Chen, T., Kornblith, S., Norouzi, M., Hinton, G.: A simple framework for
  contrastive learning of visual representations. arXiv preprint
  arXiv:2002.05709  (2020)

\bibitem{chen2019drop}
Chen, Y., Fan, H., Xu, B., Yan, Z., Kalantidis, Y., Rohrbach, M., Yan, S.,
  Feng, J.: Drop an octave: Reducing spatial redundancy in convolutional neural
  networks with octave convolution. In: Proceedings of the IEEE International
  Conference on Computer Vision. pp. 3435--3444 (2019)

\bibitem{chen2019graph}
Chen, Y., Rohrbach, M., Yan, Z., Shuicheng, Y., Feng, J., Kalantidis, Y.:
  Graph-based global reasoning networks. In: Proceedings of the IEEE Conference
  on Computer Vision and Pattern Recognition. pp. 433--442 (2019)

\bibitem{choi2019can}
Choi, J., Gao, C., Messou, J.C., Huang, J.B.: Why can't i dance in the mall?
  learning to mitigate scene bias in action recognition. In: Advances in Neural
  Information Processing Systems. pp. 853--865 (2019)

\bibitem{autoaugment}
Cubuk, E.D., Zoph, B., Mane, D., Vasudevan, V., Le, Q.V.: Autoaugment: Learning
  augmentation strategies from data. In: Proceedings of the IEEE conference on
  computer vision and pattern recognition. pp. 113--123 (2019)

\bibitem{randaugment}
Cubuk, E.D., Zoph, B., Shlens, J., Le, Q.V.: Randaugment: Practical automated
  data augmentation with a reduced search space. In: Proceedings of the
  IEEE/CVF Conference on Computer Vision and Pattern Recognition Workshops. pp.
  702--703 (2020)

\bibitem{cutout}
DeVries, T., Taylor, G.W.: Improved regularization of convolutional neural
  networks with cutout. arXiv preprint arXiv:1708.04552  (2017)

\bibitem{feichtenhofer2019slowfast}
Feichtenhofer, C., Fan, H., Malik, J., He, K.: Slowfast networks for video
  recognition. In: Proceedings of the IEEE international conference on computer
  vision. pp. 6202--6211 (2019)

\bibitem{shakeshake}
Gastaldi, X.: Shake-shake regularization. arXiv preprint arXiv:1705.07485
  (2017)

\bibitem{dropblock}
Ghiasi, G., Lin, T.Y., Le, Q.V.: Dropblock: A regularization method for
  convolutional networks. In: Advances in Neural Information Processing
  Systems. pp. 10727--10737 (2018)

\bibitem{cosinewarmup}
Goyal, P., Doll{\'a}r, P., Girshick, R., Noordhuis, P., Wesolowski, L., Kyrola,
  A., Tulloch, A., Jia, Y., He, K.: Accurate, large minibatch sgd: Training
  imagenet in 1 hour. arXiv preprint arXiv:1706.02677  (2017)

\bibitem{hara2018can}
Hara, K., Kataoka, H., Satoh, Y.: Can spatiotemporal 3d cnns retrace the
  history of 2d cnns and imagenet? In: Proceedings of the IEEE conference on
  Computer Vision and Pattern Recognition. pp. 6546--6555 (2018)

\bibitem{moco}
He, K., Fan, H., Wu, Y., Xie, S., Girshick, R.: Momentum contrast for
  unsupervised visual representation learning. In: Proceedings of the IEEE/CVF
  Conference on Computer Vision and Pattern Recognition. pp. 9729--9738 (2020)

\bibitem{robustness}
Hendrycks, D., Dietterich, T.: Benchmarking neural network robustness to common
  corruptions and perturbations. Proceedings of the International Conference on
  Learning Representations  (2019)

\bibitem{augmix}
Hendrycks, D., Mu, N., Cubuk, E.D., Zoph, B., Gilmer, J., Lakshminarayanan, B.:
  Augmix: A simple data processing method to improve robustness and
  uncertainty. arXiv preprint arXiv:1912.02781  (2019)

\bibitem{stochasticdepth}
Huang, G., Sun, Y., Liu, Z., Sedra, D., Weinberger, K.Q.: Deep networks with
  stochastic depth. In: European conference on computer vision. pp. 646--661.
  Springer (2016)

\bibitem{ji2019learning}
Ji, J., Cao, K., Niebles, J.C.: Learning temporal action proposals with fewer
  labels. In: Proceedings of the IEEE International Conference on Computer
  Vision. pp. 7073--7082 (2019)

\bibitem{dagan}
Karras, T., Aittala, M., Hellsten, J., Laine, S., Lehtinen, J., Aila, T.:
  Training generative adversarial networks with limited data. arXiv preprint
  arXiv:2006.06676  (2020)

\bibitem{regvideo}
Kim, J., Cha, S., Wee, D., Bae, S., Kim, J.: Regularization on
  spatio-temporally smoothed feature for action recognition. In: Proceedings of
  the IEEE/CVF Conference on Computer Vision and Pattern Recognition. pp.
  12103--12112 (2020)

\bibitem{adam}
Kingma, D.P., Ba, J.: Adam: A method for stochastic optimization. arXiv
  preprint arXiv:1412.6980  (2014)

\bibitem{krizhevsky2012imagenet}
Krizhevsky, A., Sutskever, I., Hinton, G.E.: Imagenet classification with deep
  convolutional neural networks. In: Advances in neural information processing
  systems. pp. 1097--1105 (2012)

\bibitem{krizhevsky2012alexnet}
Krizhevsky, A., Sutskever, I., Hinton, G.E.: Imagenet classification with deep
  convolutional neural networks. In: Advances in neural information processing
  systems. pp. 1097--1105 (2012)

\bibitem{kuehne2011hmdb}
Kuehne, H., Jhuang, H., Garrote, E., Poggio, T., Serre, T.: Hmdb: a large video
  database for human motion recognition. In: 2011 International Conference on
  Computer Vision. pp. 2556--2563. IEEE (2011)

\bibitem{smoothmix}
Lee, J.H., Zaigham~Zaheer, M., Astrid, M., Lee, S.I.: Smoothmix: A simple yet
  effective data augmentation to train robust classifiers. In: Proceedings of
  the IEEE/CVF Conference on Computer Vision and Pattern Recognition Workshops.
  pp. 756--757 (2020)

\bibitem{attributemix}
Li, H., Zhang, X., Xiong, H., Tian, Q.: Attribute mix: Semantic data
  augmentation for fine grained recognition. arXiv preprint arXiv:2004.02684
  (2020)

\bibitem{li2018resound}
Li, Y., Li, Y., Vasconcelos, N.: Resound: Towards action recognition without
  representation bias. In: Proceedings of the European Conference on Computer
  Vision (ECCV). pp. 513--528 (2018)

\bibitem{fastautoaugment}
Lim, S., Kim, I., Kim, T., Kim, C., Kim, S.: Fast autoaugment. In: Advances in
  Neural Information Processing Systems. pp. 6665--6675 (2019)

\bibitem{lin2019tsm}
Lin, J., Gan, C., Han, S.: Tsm: Temporal shift module for efficient video
  understanding. In: Proceedings of the IEEE International Conference on
  Computer Vision. pp. 7083--7093 (2019)

\bibitem{cosinelr}
Loshchilov, I., Hutter, F.: Sgdr: Stochastic gradient descent with warm
  restarts. arXiv preprint arXiv:1608.03983  (2016)

\bibitem{luo2020viprior}
Luo, Z., Xu, D., Zhang, Z.: Challenge report:vipriors action recognition
  challenge. arXiv preprint arXiv:2007.08180  (2020)

\bibitem{pirl}
Misra, I., Maaten, L.v.d.: Self-supervised learning of pretext-invariant
  representations. In: Proceedings of the IEEE/CVF Conference on Computer
  Vision and Pattern Recognition. pp. 6707--6717 (2020)

\bibitem{ryoo2019assemblenet}
Ryoo, M.S., Piergiovanni, A., Tan, M., Angelova, A.: Assemblenet: Searching for
  multi-stream neural connectivity in video architectures. arXiv preprint
  arXiv:1905.13209  (2019)

\bibitem{vggnet}
Simonyan, K., Zisserman, A.: Very deep convolutional networks for large-scale
  image recognition. In: International Conference on Learning Representations
  (2015)

\bibitem{hideandseek}
Singh, K.K., Lee, Y.J.: Hide-and-seek: Forcing a network to be meticulous for
  weakly-supervised object and action localization. In: 2017 IEEE international
  conference on computer vision (ICCV). pp. 3544--3553. IEEE (2017)

\bibitem{fixmatch}
Sohn, K., Berthelot, D., Li, C.L., Zhang, Z., Carlini, N., Cubuk, E.D.,
  Kurakin, A., Zhang, H., Raffel, C.: Fixmatch: Simplifying semi-supervised
  learning with consistency and confidence. arXiv preprint arXiv:2001.07685
  (2020)

\bibitem{soomro2012ucf101}
Soomro, K., Zamir, A.R., Shah, M.: Ucf101: A dataset of 101 human actions
  classes from videos in the wild. arXiv preprint arXiv:1212.0402  (2012)

\bibitem{dropout}
Srivastava, N., Hinton, G., Krizhevsky, A., Sutskever, I., Salakhutdinov, R.:
  Dropout: a simple way to prevent neural networks from overfitting. The
  journal of machine learning research  \textbf{15}(1),  1929--1958 (2014)

\bibitem{stroud2020d3d}
Stroud, J., Ross, D., Sun, C., Deng, J., Sukthankar, R.: D3d: Distilled 3d
  networks for video action recognition. In: The IEEE Winter Conference on
  Applications of Computer Vision. pp. 625--634 (2020)

\bibitem{tran2015learning}
Tran, D., Bourdev, L., Fergus, R., Torresani, L., Paluri, M.: Learning
  spatiotemporal features with 3d convolutional networks. In: Proceedings of
  the IEEE international conference on computer vision. pp. 4489--4497 (2015)

\bibitem{tran2018closer}
Tran, D., Wang, H., Torresani, L., Ray, J., LeCun, Y., Paluri, M.: A closer
  look at spatiotemporal convolutions for action recognition. In: Proceedings
  of the IEEE conference on Computer Vision and Pattern Recognition. pp.
  6450--6459 (2018)

\bibitem{manimixup}
Verma, V., Lamb, A., Beckham, C., Najafi, A., Mitliagkas, I., Lopez-Paz, D.,
  Bengio, Y.: Manifold mixup: Better representations by interpolating hidden
  states. In: International Conference on Machine Learning. pp. 6438--6447
  (2019)

\bibitem{attentivecutmix}
Walawalkar, D., Shen, Z., Liu, Z., Savvides, M.: Attentive cutmix: An enhanced
  data augmentation approach for deep learning based image classification. In:
  ICASSP 2020-2020 IEEE International Conference on Acoustics, Speech and
  Signal Processing (ICASSP). pp. 3642--3646. IEEE (2020)

\bibitem{wang2016temporal}
Wang, L., Xiong, Y., Wang, Z., Qiao, Y., Lin, D., Tang, X., Van~Gool, L.:
  Temporal segment networks: Towards good practices for deep action
  recognition. In: European conference on computer vision. pp. 20--36. Springer
  (2016)

\bibitem{wang2018non}
Wang, X., Girshick, R., Gupta, A., He, K.: Non-local neural networks. In:
  Proceedings of the IEEE conference on computer vision and pattern
  recognition. pp. 7794--7803 (2018)

\bibitem{uda}
Xie, Q., Dai, Z., Hovy, E., Luong, M.T., Le, Q.V.: Unsupervised data
  augmentation for consistency training. arXiv preprint arXiv:1904.12848
  (2019)

\bibitem{noisystudent}
Xie, Q., Luong, M.T., Hovy, E., Le, Q.V.: Self-training with noisy student
  improves imagenet classification. In: Proceedings of the IEEE/CVF Conference
  on Computer Vision and Pattern Recognition. pp. 10687--10698 (2020)

\bibitem{xie2017rethinking}
Xie, S., Sun, C., Huang, J., Tu, Z., Murphy, K.: Rethinking spatiotemporal
  feature learning for video understanding. arXiv preprint arXiv:1712.04851
  \textbf{1}(2), ~5 (2017)

\bibitem{shakedrop}
Yamada, Y., Iwamura, M., Akiba, T., Kise, K.: Shakedrop regularization for deep
  residual learning. IEEE Access  \textbf{7},  186126--186136 (2019)

\bibitem{cutblur}
Yoo, J., Ahn, N., Sohn, K.A.: Rethinking data augmentation for image
  super-resolution: A comprehensive analysis and a new strategy. In:
  Proceedings of the IEEE/CVF Conference on Computer Vision and Pattern
  Recognition. pp. 8375--8384 (2020)

\bibitem{cutmix}
Yun, S., Han, D., Oh, S.J., Chun, S., Choe, J., Yoo, Y.: Cutmix: Regularization
  strategy to train strong classifiers with localizable features. In:
  Proceedings of the IEEE International Conference on Computer Vision. pp.
  6023--6032 (2019)

\bibitem{crgan}
Zhang, H., Zhang, Z., Odena, A., Lee, H.: Consistency regularization for
  generative adversarial networks. In: International Conference on Learning
  Representations (2019)

\bibitem{mixup}
Zhang, H., Cisse, M., Dauphin, Y.N., Lopez-Paz, D.: mixup: Beyond empirical
  risk minimization. arXiv preprint arXiv:1710.09412  (2017)

\bibitem{diffauggan}
Zhao, S., Liu, Z., Lin, J., Zhu, J.Y., Han, S.: Differentiable augmentation for
  data-efficient gan training. arXiv preprint arXiv:2006.10738  (2020)

\bibitem{bcrgan}
Zhao, Z., Singh, S., Lee, H., Zhang, Z., Odena, A., Zhang, H.: Improved
  consistency regularization for gans. arXiv preprint arXiv:2002.04724  (2020)

\bibitem{cam}
Zhou, B., Khosla, A., Lapedriza, A., Oliva, A., Torralba, A.: Learning deep
  features for discriminative localization. In: Proceedings of the IEEE
  conference on computer vision and pattern recognition. pp. 2921--2929 (2016)

\end{thebibliography}

\end{document}